\documentclass[letterpaper]{article} % DO NOT CHANGE THIS
\usepackage[preprint]{aaai2027}  % arXiv/preprint version
\usepackage[hyphens]{url}  % DO NOT CHANGE THIS
\usepackage{graphicx} % DO NOT CHANGE THIS
\usepackage{natbib}  % DO NOT CHANGE THIS AND DO NOT ADD ANY OPTIONS TO IT
\usepackage{caption} % DO NOT CHANGE THIS AND DO NOT ADD ANY OPTIONS TO IT
\usepackage{algorithm}
\usepackage{algorithmic}
\usepackage{booktabs}
\usepackage{array}
\title{GrapNet: A Programmable Dynamic-Architecture Neural Graph Substrate}
\author{Zirong Li}
\affiliations{Tiangong University\\
\texttt{xingyeyouyv@gmail.com}}

\begin{document}
\maketitle

\begin{abstract}
Programmability is a missing first-class interface in fixed-tensor neural networks: editing a relation, freezing a subgraph, auditing a local function, or changing the execution backend should be an operation on the neural program rather than ad-hoc parameter surgery. GrapNet studies this graph-as-network setting. The graph is the architecture and executable program, not an input data graph. Each current-layer compute node owns its next-layer child references and a trainable allocation vector aligned with those references; deleting a relation physically removes both the child reference and the corresponding allocation coordinate. Structural rules and execution policies live outside the node core, so the same child-owned graph can be grown, frozen, structurally edited, routed by attention over active relations, grouped into trainable family blocks, or lowered to dense snapshots after topology stabilizes. GrapNet composes with conventional modules through a vector-valued parent interface: dense layers, CNN encoders, ResNet feature extractors, attention blocks, and transformer representations can all feed one sensory GrapNode per coordinate, while MLPs and Transformer FFNs are fully connected GrapNet special cases. The evaluation is organized as a programmability stress suite rather than as a new replay benchmark: relation-local storage invariants, gradient equivalence, physical deletion, family and dense execution controllers, attention-policy gradients, Transformer-FFN replacement, tabular covariance editing, and local structural intervention are tested over the same canonical graph. Continual-learning streams are used as one demanding setting because they require online growth, replay, freezing choices, and task-free calibration to coexist. In a matched ten-seed Split Fashion-MNIST study, a plastic GrapNet+ER head reaches 63.16\% seen-class accuracy versus 51.08\% for a parameter-larger dense MLP+ER under the same seen-class loss and replay memory (paired $\Delta=12.08$ points, $p=1.3{\times}10^{-5}$). On Split CIFAR-10 with a frozen ImageNet ResNet-18 encoder, the same substrate improves the online head over MLP-256 by 3.81 points ($p=0.0026$). These results support GrapNet as an editable neural graph substrate whose core value is structural programmability with faithful execution views.
\end{abstract}

\section{Introduction}
GNNs treat a graph as an input object and learn shared neural functions over its nodes and edges. GrapNet treats a graph as the neural network itself. This dynamic-architecture graph neural network setting needs a different ownership invariant: topology edits must change the executable program, remain visible to training rules, and still admit efficient tensor backends. Dense tensors are an excellent execution format and a poor ownership model for that setting. A masked weight can be disabled, but the rectangular slot, optimizer state, and global tensor layout still remain. A hand-edited module can add or remove Python objects, but relation semantics then spread across module lists, parameter surgery, optimizer refreshes, and execution special cases.

GrapNet addresses this mismatch by storing the network as a child-owned computation graph. For a transition from layer $l$ to layer $l+1$, every current-layer GrapNode owns a list of next-layer child references and a trainable allocation vector $W_{alloc}$ aligned with that list:
\[
\texttt{node.children}[k] \leftrightarrow \texttt{node.W\_alloc}[0,k].
\]
The relation is the paired child reference and allocation coordinate. Removing slot $k$ removes both. Adding a relation appends both. Dense, sparse, family-block, and compiled tensors are execution views reconstructed from this canonical object.

GrapNet composes with conventional neural modules rather than replacing them. It does not require raw pixels or hand-designed scalar inputs. Any vector-producing parent interface can feed the sensory layer: a fully connected layer, a convolutional encoder, a frozen ResNet feature extractor, a pooled transformer representation, or an attention module output. The vector coordinates are read in parallel by sensory GrapNodes, one coordinate per sensory node. The relationship is stronger for MLPs: a dense MLP is recovered exactly by choosing a fully connected child-owned topology. Thus the position-wise FFN inside a Transformer block can be instantiated as a GrapNet block while self-attention remains the token-mixing operator. The contribution is the editable child-owned relation structure exposed at or after these vector interfaces.

The present paper positions structural editability at the decision head, where online task conflicts, replay calibration, and local structural interventions accumulate. Whether top-down structural control from an editable head can regulate a sensory encoder is compatible with the substrate, but is a separate question beyond the scope of this work.

The paper makes three claims. First, child ownership gives neural topology a stable edit invariant: relation deletion, node deletion, optimizer refresh, and downstream continuity become explicit operations on the graph. Second, placing structural rules outside the GrapNode core makes growth, freezing, pruning, covariance sifting, and replay controllers composable without changing node semantics. Third, execution can follow topology state: the editable graph trains through live family-vectorized blocks while stable graphs can be lowered to dense or compiled endpoints.

The contribution is therefore not a new pruning score, a task router, or a new replay buffer. It is a representation boundary: the graph object stores the editable neural program, and policies decide how that program is grown, frozen, simplified, routed, and executed. The appropriate comparison is not merely whether a fixed dense layer is faster when no edit is required; it is whether relation insertion, physical deletion, optimizer-aware refresh, programmable freezing, local intervention, and backend rebuilding exist as stable operations rather than fragile tensor surgery. Dense layers and pretrained heads remain the right throughput baseline when topology is fixed, but they do not provide these edit operations as first-class interfaces. The experiments use continual and online learning as stress tests for this substrate because they force programmable edits to coexist with ordinary gradient training. We therefore use matched replay streams, seen-class losses, parameter accounting, and head-only visual-feature comparisons while also reporting non-image covariance editing, relation-cut localization, attention execution, Transformer-FFN replacement, and dense-snapshot lowering.

The empirical section is organized around those claims. The main online continual-learning comparison uses a matched seen-class protocol: dense and GrapNet heads see the same Split Fashion-MNIST stream, the same replay memory, and the same cross-entropy over classes seen so far. Under this protocol, a plastic GrapNet head improves global seen-class calibration over dense replay heads with more parameters, while preserving task-local accuracy. A frozen GrapNet variant locks previous hidden nodes and exposes the expected trade-off: lower structural drift and lower global plasticity. A second study uses a frozen ImageNet ResNet-18 encoder on Split CIFAR-10 and trains only the online head, verifying the general vector-interface claim for conventional visual features. A non-image tabular study keeps one sensory node per scalar feature and uses covariance to prune deep shared subgraphs, delete hidden nodes, and rebuild execution families. A structural localization audit then cuts task-owned child relations and measures the resulting task-specific accuracy drop. Execution experiments show that trainable family blocks are a faithful dynamic backend for the same canonical graph.

\section{Related Work}
\textbf{Editable and sparse neural structure.}
Pruning masks, lottery-ticket subnetworks, dynamic sparse training, and PackNet-style task allocation operate over tensorized models \citep{mallya2018packnet,frankle2019lottery,evci2020rigging,lasby2023srigl}. GrapNet starts from the opposite side: the editable node-owned relation is canonical, and tensor forms are backend choices. The distinction matters for structural rules because a removed child relation is gone from storage, gradient flow, and subsequent execution views.

\textbf{Continual learning and dynamic architecture.}
Continual learning methods use regularization, replay, expansion, masking, distillation, and routing to preserve old competence \citep{kirkpatrick2017overcoming,zenke2017continual,rusu2016progressive,yoon2018den,serra2018hat,chaudhry2019tiny,buzzega2020dark,prabhu2020gdumb,boschini2023xder}. Architecture itself is now recognized as an active variable in continual learning \citep{mirzadeh2022architecture,douillard2022dytox,wang2022l2p,wang2024survey}. GrapNet contributes a representation substrate for those variables: growth and freezing are external structural policies over child-owned nodes, while replay remains an external sample-memory controller.

\textbf{Graph computation and ML systems.}
PyG and DGL implement \emph{graphs as data}: the graph is an input object whose nodes and edges carry features, while learned message functions are shared over that input graph \citep{fey2019fast,wang2019dgl}. GrapNet implements \emph{graphs as network}: the graph is the neural program whose nodes own parameters, activations, child references, and execution-relevant relation slots. The same word ``graph'' therefore denotes different abstraction levels. GNN libraries are natural backends or modeling tools for data graphs; GrapNet addresses the storage and edit semantics of the model graph itself.

Attention mechanisms use input-dependent routing as a core computational pattern \citep{vaswani2017attention}. GrapNet realizes this pattern as an execution policy over active child-owned relations rather than as a replacement for canonical topology. Modern ML systems also separate flexible programs from optimized execution \citep{paszke2019pytorch,ansel2024pytorch2}. GrapNet applies the separation inside the neural topology: child-owned storage is canonical, and execution policies lower it to scatter, attention-gated relations, trainable family blocks, dense snapshots, or compiled relation kernels.

\section{Child-Owned Computation}
\subsection{Canonical Relation Ownership}
A GrapNode stores local state $(b_u,\phi_u)$, next-layer child references $C_u=[v_1,\ldots,v_{d_u}]$, and an aligned allocation vector $W^u_{alloc}=[w_1,\ldots,w_{d_u}]$. The invariant is
\[
I(u):\quad |C_u| = |W^u_{alloc}|,\qquad C_u[k] \leftrightarrow W^u_{alloc}[k].
\]
Relation deletion at local slot $k$ maps
\[
(C_u,W^u_{alloc}) \mapsto (C_u\setminus C_u[k],\; W^u_{alloc}\setminus W^u_{alloc}[k]).
\]
Node deletion is a separate graph operation. This keeps pointer deletion, real node deletion, and downstream continuity distinguishable. The invariant also fixes the optimizer-facing semantics. When a relation is added, a new allocation coordinate is a new trainable parameter slot and the optimizer can be refreshed explicitly. When a relation is removed, its coordinate disappears from the canonical parameter set rather than remaining as a masked entry whose state must be tracked separately.

\paragraph{Proposition 1 (storage invariant and edit scope).}
Let $R(u)=\{(C_u[k],W^u_{alloc}[k])\}_{k=1}^{d_u}$ be the outgoing relation multiset owned by node $u$. A legal relation edit rewrites only $R(u)$, followed by optimizer refresh and execution-view rebuild. Its logical storage scope is $O(d_u)$ rather than the size of the materialized layer matrix. \emph{Proof sketch.} The only canonical objects modified by deleting local slot $k$ are the reference $C_u[k]$ and the aligned allocation coordinate $W^u_{alloc}[k]$; all other nodes keep their relation lists and allocation coordinates. Adding a relation appends one child reference and one trainable coordinate to the same local pair. Therefore the edit is local in the degree of $u$. A mask over a dense matrix can disable a coordinate but leaves the rectangular slot and its optimizer state in the canonical tensor layout. Physical dense parameter surgery can remove a row or column, but must then migrate global tensor indices, optimizer state, downstream module references, and execution kernels. GrapNet stores the relation as the canonical object, so storage, gradient participation, and execution views agree after the edit.

\paragraph{Proposition 2 (materialized-matrix and gradient equivalence).}
For a fixed child-owned topology with active relation set $E$, define a materialized matrix $A$ by $A_{ij}=w_{ij}$ if $(u_i,v_j)\in E$ and $A_{ij}=0$ otherwise, and let $c_j=b_j$. The child-owned transition and the matrix transition compute the same pre-activation for every sample:
\[
\tilde z^{(b)}_j=\sum_{i:(u_i,v_j)\in E}h_i^{(b)}w_{ij}+b_j
=\sum_i h_i^{(b)}A_{ij}+c_j=z^{(b)}_j.
\]
Consequently, for any differentiable downstream loss, reverse-mode gradients on active relations are identical:
\[
\frac{\partial\mathcal L}{\partial w_{ij}}=\sum_b\frac{\partial\mathcal L}{\partial z_j^{(b)}}h_i^{(b)},\quad
\frac{\partial\mathcal L}{\partial b_j}=\sum_b\frac{\partial\mathcal L}{\partial z_j^{(b)}}.
\]
\emph{Proof sketch.} The two transitions produce identical $z_j^{(b)}$ values because inactive relations are absent in the child-owned sum and zero in $A$. The same scalar computation graph is therefore evaluated up to storage layout. Differentiating the shared scalar expression gives the displayed local gradients. Execution policies such as reference scatter, trainable family blocks, and dense snapshots preserve these gradients when they assemble tensors from live $W_{alloc}$ Parameters without detaching them; they change evaluation layout, not the learning rule.

\begin{figure*}[t]
\centering
\includegraphics[width=0.86\textwidth]{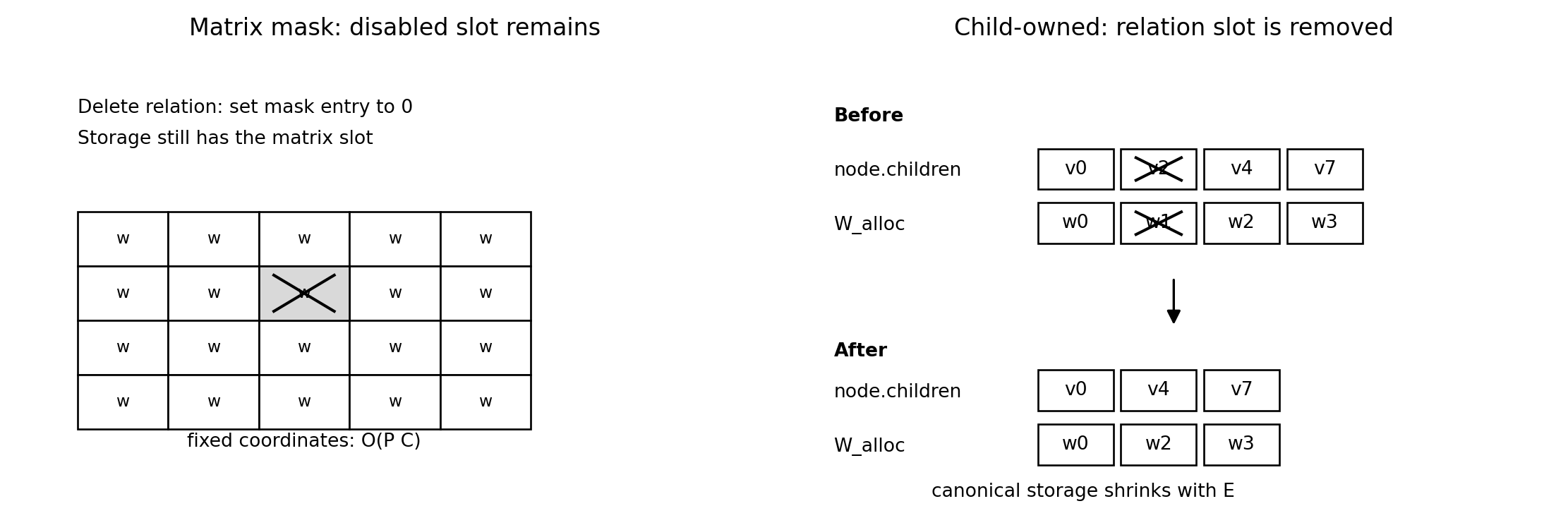}
\caption{Child-owned deletion versus matrix masking. A masked dense matrix leaves a disabled relation as a fixed coordinate in the rectangular parameter layout. GrapNet removes both the next-layer reference $C_u[k]$ and the aligned allocation coordinate $W^u_{alloc}[k]$ from the canonical node-owned structure. Tensor views are rebuilt from the remaining physical relations.}
\label{fig:invariant}
\end{figure*}

\subsection{MLPs and Transformer FFNs as Special Cases}
The fully connected MLP case follows by direct parameter mapping. Consider a dense layer with input $h\in\mathbf{R}^{n}$, weight matrix $A\in\mathbf{R}^{n\times m}$, bias $c\in\mathbf{R}^{m}$, and output activation $\phi_j$:
\[
q_j = \sum_{i=1}^{n} h_i A_{ij} + c_j,\qquad h'_j=\phi_j(q_j).
\]
Instantiate current-layer GrapNodes $u_1,\ldots,u_n$ and next-layer GrapNodes $v_1,\ldots,v_m$. Set
\[
C_{u_i}=[v_1,\ldots,v_m],\quad W^{u_i}_{alloc}[j]=A_{ij},\quad b_{v_j}=c_j.
\]
The GrapNet transition then gives
\[
\tilde q_j=\sum_{i:v_j\in C_{u_i}} h_i W^{u_i}_{alloc}[j] + b_{v_j}
       =\sum_{i=1}^{n} h_i A_{ij}+c_j=q_j.
\]
Applying the same mapping layer by layer proves that every feed-forward MLP is a GrapNet topology with fully connected child lists. The released equivalence audit copies PyTorch dense-layer weights into $W_{alloc}$ and biases and reports maximum absolute differences up to $1.2\times 10^{-7}$ across linear, ReLU, and GELU MLP settings.

This also covers the Transformer FFN sublayer. For token states $H\in\mathbf{R}^{B\times T\times D}$, a position-wise FFN applies the same MLP $F$ to each token: $Y_{bt}=F(H_{bt})$. Flattening $H$ to $\bar H\in\mathbf{R}^{(BT)\times D}$, applying the equivalent GrapNet implementation of $F$, and reshaping back to $B\times T\times D'$ gives the same FFN values. Self-attention, residual connections, and normalization remain token-mixing and stabilization operators; GrapNet supplies an editable child-owned implementation of the MLP/FFN transformation. A supplemental Tiny Transformer audit copies a standard FFN into GrapNet-FFN and reports FFN forward error $1.2\times 10^{-7}$, Transformer-logit error $6.0\times 10^{-8}$, matched FFN gradients below $7.5\times 10^{-9}$, identical 93.36\% test accuracy after matched training, and physical FFN relation deletion from 1024 to 992 active relations.

\subsection{Forward Semantics}
For batch activations $X\in\mathbf{R}^{B\times P}$ and a transition with $E$ active child-owned relations, the reference backend computes
\[
Z[b,t(e)] \mathrel{+}= X[b,s(e)]\,w_e,
\]
where source $s(e)$ and target $t(e)$ are derived from node-owned child lists. Algebraically this is a sparse linear map $Z=XW+b$, but $W$ is an execution view. The canonical object remains the node-owned relation set.

\begin{algorithm}[t]
\caption{Child-Owned Transition}
\begin{algorithmic}[1]
\STATE Input activations $X$, current nodes $U$, next nodes $V$
\STATE Initialize next buffer $Z$ to zero
\FOR{each current-layer node $u_i$ in $U$}
  \STATE Read child list $C_i$ and allocation vector $W^i_{alloc}$
  \STATE Compute local messages $M_i=X[:,i]W^i_{alloc}$
  \FOR{each local child slot $k$ in $C_i$}
    \STATE Add $M_i[:,k]$ into child $C_i[k]$
  \ENDFOR
\ENDFOR
\STATE Apply next-node bias and activation functions
\end{algorithmic}
\end{algorithm}

\subsection{Rules and Execution Policies}
The GrapNode core contains structural fields and local computation. Growth, freezing, pruning, replay control, covariance sifting, and backend selection are external policies. Structural rule managers can inject, replace, enable, and disable rules at runtime. Execution policy managers redirect selected transitions to alternative backends while preserving the callable graph.

Trainable family execution is the dynamic fast path. Current-layer nodes with identical next-layer child-reference patterns form an execution family. For family $f$, member activations form $H_f\in\mathbf{R}^{B\times m_f}$ and live allocation parameters form $W_f\in\mathbf{R}^{m_f\times k_f}$. The backend evaluates
\[
M_f = H_f W_f
\]
and scatters $M_f$ to the corresponding next-layer targets. A family is not canonical storage: it is metadata built from the current child lists. During a topology-stable interval, the metadata is reused, while each forward pass assembles $W_f$ from live $W_{alloc}$ parameters. Backpropagation therefore follows the differentiable path $M_f\leftarrow W_f\leftarrow W_{alloc}$ and accumulates gradients in the original child-owned parameters; no detached family weights and no manual post-hoc gradient scattering are used.

Structural edits are applied only between optimization steps. A training step completes forward, loss evaluation, backward, and optimizer update under one topology version. If a rule then adds or removes relations or nodes, the canonical child lists and aligned $W_{alloc}$ vectors are rebuilt, the optimizer is refreshed over current parameters, and the family controller rebuilds its grouping before the next forward pass. Deleted relations are absent from later computation graphs and receive no later gradients; retained relations remain live parameters. Stable topology can then be lowered by an external dense-snapshot controller, sibling to the freezing and family controllers, which packs the current child-owned graph into fixed dense transition matrices for inference or profiling and is rebuilt after later edits.

Attention is another execution policy over the same relation set. For active relation $(i,j)$, an attention backend can compute
\[
a_{bij}=\mathrm{softmax}_{i:(i,j)\in E}(h_{bi}k_iq_j)
\]
and
\[
z_{bj}=\sum_{i:(i,j)\in E} a_{bij}h_{bi}w_{ij},
\]
where $E$ is read from current child lists and $w_{ij}$ is the child-owned allocation coordinate. A structural edit changes both the executable relation set and the attention mask.

These policies are intended to be native drivers rather than post-hoc repairs. The reference scatter path is useful for correctness and debugging; family execution is the dynamic acceleration path for repeated child patterns during online editing; attention execution adds input-dependent weighting over the same relation set; dense snapshots and compiled kernels are throughput endpoints after topology stabilizes. The dense snapshot endpoint is not a second model: it is an external controller that materializes the current relation set as fixed matrices and is refreshed when the graph is edited. All execution policies read from the same child-owned graph.

\paragraph{Objective and gradient path.}
GrapNet changes the parameterization of the head, not the learning objective. In the online class-incremental studies, the loss at task $t$ is the ordinary seen-class replay objective
\[
\mathcal{L}_t=\mathrm{CE}(g(x_t)_{S_t},y_t)+\lambda\,\mathrm{E}_{(x,y)\sim\mathcal{M}}\mathrm{CE}(g(x)_{S_t},y),
\]
where $S_t$ is the set of classes observed so far, $\mathcal{M}$ is the replay buffer, and $g$ is either a dense head or a GrapNet head. For a child-owned relation from parent $i$ to child $j$, node message passing is still an ordinary differentiable computation:
\[
\begin{array}{l}
m^{(b)}_{i\rightarrow j}=h_i^{(b)}w_{ij},\qquad
z_j^{(b)}=\sum_{i:u_i\rightarrow v_j}m^{(b)}_{i\rightarrow j}+b_j,\\
h_j'^{(b)}=\phi_j(z_j^{(b)}).
\end{array}
\]
The child-owned edge coordinate and the node-owned bias therefore receive the standard dense-layer local gradients
\[
\begin{array}{l}
\displaystyle
\frac{\partial\mathcal{L}}{\partial w_{ij}}
=\sum_b\frac{\partial\mathcal{L}}{\partial z_j^{(b)}}h_i^{(b)},\qquad
\displaystyle
\frac{\partial\mathcal{L}}{\partial b_j}
=\sum_b\frac{\partial\mathcal{L}}{\partial z_j^{(b)}}.
\end{array}
\]
Both $w_{ij}$ inside $W^{u_i}_{alloc}$ and $b_j$ are live PyTorch Parameters. GrapNet therefore introduces no custom backward rule: autograd propagates through multiplication, summation, bias addition, and activation as in dense layers, while storage remains node-owned rather than matrix-slot-owned. Family execution preserves this gradient path because each family block is assembled from live allocation coordinates under the current topology version.

\section{Programmability Stress Tests}
All reported studies use released scripts and JSON outputs. Online studies use one pass over five binary tasks $(0,1),(2,3),(4,5),(6,7),(8,9)$, replay memory 500 unless stated otherwise, and two metrics: task-local pair accuracy and global seen-class accuracy. The global metric takes an argmax over classes seen so far using a task-free evaluation rule. The matched studies train dense and GrapNet heads with the same cross-entropy over seen classes for both current and replay batches.

The stress suite is ordered by programmability first and benchmark performance second. Structural localization and covariance editing test what fixed dense heads do not expose as stable operations: physical relation deletion, local intervention, optimizer-aware rebuilding, and execution-view refresh. The matched continual-learning studies then test whether the same substrate remains competitive under ordinary seen-class training. To keep the CL results from becoming a method-paper confound, no task identity is used for seen-class evaluation; dense and GrapNet runs share the same task order, minibatch stream, replay insertion rule, replay samples, optimizer family, seeds, and seen-class CE; and parameter counts are reported. Frozen variants are policy ablations: freezing tests structural memory, while plastic replay tests global calibration under the same child-owned substrate.

\paragraph{Stress-suite order.}
We foreground programmable structure before CL accuracy: relation-cut localization tests causal addressability of owned relations; tabular covariance editing tests physical relation and node deletion outside image features; matched Split Fashion-MNIST and CIFAR-10 ResNet-feature heads test ordinary seen-class learning under the same substrate; family, attention, dense-snapshot, and Transformer-FFN audits test execution-policy replacement over the same canonical graph.

\subsection{Structural Localization by Relation Cut}
The first audit tests the core programmability claim before reporting CL accuracy: learned structure should remain physically addressable after training. On the Split Fashion-MNIST task order, each task receives 48 newly added hidden nodes. Those nodes own output child relations only to the task's two classes. After training all five tasks, we copy the trained graph and cut the physical output child relations owned by one target task. The cut set is selected only from task ownership metadata; test performance is measured after the cut.

\begin{table}[t]
\centering
\scriptsize
\begin{tabular}{lccc}
\toprule
Metric & Mean & Std. & Audit \\
\midrule
Base pair accuracy & 98.26 & 0.25 & 10 seeds \\
Target-pair drop & 48.26 & 0.25 & 96 cuts/task \\
Non-target-pair drop & 0.00 & 0.00 & 4 tasks/cut \\
Localization margin & 48.26 & 0.25 & $p=6.8{\times}10^{-22}$ \\
\bottomrule
\end{tabular}
\caption{Structural localization on Split Fashion-MNIST. Cutting the task-owned child relations selectively removes the owning task's computation while preserving non-target task performance. Every run removed exactly the expected child-reference/allocation-slot pairs.}
\label{tab:localization}
\end{table}

Table~\ref{tab:localization} shows the expected graph-as-network behavior. Relation cuts are exact storage edits: every target task removes $48{\times}2$ child-reference/allocation-slot pairs. The owning task drops from near-perfect pair accuracy to chance-level behavior, while non-target tasks remain unchanged within measurement precision. This gives a direct causal readout for the child-owned architecture graph. The audit should be read as structural localization, not as a claim that GrapNet automatically discovers a semantic task graph: the ownership metadata is created by the growth policy, and the experiment verifies that those owned relations remain editable and causally meaningful after learning.

\subsection{Matched Split Fashion-MNIST}
Table~\ref{tab:fashion} reports the main ten-seed Fashion-MNIST study \citep{xiao2017fashion}. Images are downsampled to $7{\times}7$ so the study isolates online structure and replay rather than image modeling. Dense ER-256 has more trainable parameters than GrapNet-48. GrapNet plastic keeps previous hidden nodes trainable under replay; GrapNet frozen locks previous hidden nodes after each task. All rows use the same one-pass stream and the same seen-class objective, so the table compares head structure and structural policy rather than a different evaluation loss.

\begin{table*}[t]
\centering
\scriptsize
\resizebox{\textwidth}{!}{%
\begin{tabular}{lccccc}
\toprule
Method & Params & Task-local acc. & Seen-class acc. & Global forgetting & Train time \\
\midrule
Dense ER-64, reservoir & 3.85K & 90.91 $\pm$ 3.51 & 48.48 $\pm$ 3.84 & 47.86 $\pm$ 5.23 & 0.25s \\
Dense ER-256, reservoir & 15.37K & 96.45 $\pm$ 0.53 & 51.08 $\pm$ 2.51 & 52.24 $\pm$ 2.76 & 0.26s \\
Dense ER-256, balanced & 15.37K & 96.41 $\pm$ 0.55 & 50.19 $\pm$ 1.84 & 53.90 $\pm$ 2.10 & 0.27s \\
GrapNet-48 frozen, reservoir & 13.45K & 96.01 $\pm$ 0.78 & 54.47 $\pm$ 4.29 & 48.24 $\pm$ 5.13 & 5.48s \\
GrapNet-48 plastic, reservoir & 13.45K & 96.41 $\pm$ 0.77 & \textbf{63.16 $\pm$ 3.20} & \textbf{36.34 $\pm$ 4.18} & 3.57s \\
GrapNet-48 plastic, balanced & 13.45K & \textbf{96.47 $\pm$ 0.61} & 62.71 $\pm$ 2.50 & 36.75 $\pm$ 3.82 & 3.37s \\
\bottomrule
\end{tabular}%
}
\caption{Matched one-pass Split Fashion-MNIST. Dense and GrapNet rows use the same seen-class loss, replay protocol, stream, and seeds. GrapNet plastic improves seen-class accuracy over dense ER-256 reservoir by 12.08 points (95\% CI [8.89, 15.26], paired $p=1.3{\times}10^{-5}$). GrapNet times are live dynamic-training times with editable topology; family execution is the dynamic acceleration path, while dense snapshots are stabilized-topology throughput endpoints.}
\label{tab:fashion}
\end{table*}

The plastic GrapNet head improves the class-incremental metric while matching task-local accuracy: against dense ER-256 reservoir, the paired seen-class gain is 12.08 points and global forgetting drops by 15.90 points ($p=8.6{\times}10^{-6}$). Against dense ER-256 balanced replay, the gain is 12.52 points ($p=1.4{\times}10^{-6}$). The result is not explained by parameter count: GrapNet-48 reports 13.45K trainable parameters versus 15.37K for dense ER-256. It is also not explained by a task-local oracle, because the decisive metric is the task-free argmax over all classes seen so far.

The frozen-policy ablation identifies the structural-memory operating point. Freezing previous hidden nodes keeps task-local accuracy high but lowers the global seen-class metric because old hidden subgraphs cannot recalibrate logits as new classes arrive. Keeping them plastic under replay is the effective class-incremental head. GrapNet makes this distinction explicit: freezing is a structural policy over node ownership, while replay and seen-class CE remain ordinary learning policies over the same graph.

\subsection{CIFAR-10 with a Frozen ResNet Encoder}
The second study tests encoder composition on CIFAR-10 \citep{krizhevsky2009learning}. A torchvision ResNet-18 with ImageNet weights \citep{he2016deep} is frozen and maps each image to a 512-dimensional feature vector. The encoder is shared by all methods and is never updated during the online stream; only the head is trained. Thus the comparison isolates whether a child-owned head can use conventional visual features more effectively than dense heads under the same replay and seen-class loss. Table~\ref{tab:cifar} reports five seeds.

\begin{table}[t]
\centering
\scriptsize
\resizebox{\columnwidth}{!}{%
\begin{tabular}{lcccc}
\toprule
Head & Params & Task-local & Seen-class & Forgetting \\
\midrule
Linear ER & 5.13K & 94.14 $\pm$ 0.47 & 57.28 $\pm$ 2.29 & 39.85 $\pm$ 3.85 \\
MLP-256 ER & 133.90K & 94.74 $\pm$ 0.65 & 67.07 $\pm$ 2.36 & 30.66 $\pm$ 2.26 \\
GrapNet frozen & 124.57K & 94.92 $\pm$ 0.72 & 64.72 $\pm$ 3.07 & 36.02 $\pm$ 3.12 \\
GrapNet plastic & 124.57K & \textbf{95.14 $\pm$ 0.43} & \textbf{70.88 $\pm$ 1.72} & \textbf{28.26 $\pm$ 2.74} \\
\bottomrule
\end{tabular}%
}
\caption{Split CIFAR-10 using a frozen ImageNet ResNet-18 feature encoder. Online heads use matched seen-class CE and reservoir replay. GrapNet plastic improves seen-class accuracy over MLP-256 by 3.81 points (95\% CI [2.24, 5.38], paired $p=0.0026$).}
\label{tab:cifar}
\end{table}

The CIFAR result confirms the vector-valued sensory interface. Each encoder coordinate feeds one GrapNet sensory node, and child-owned growth builds an editable head on top. The plastic GrapNet head improves seen-class accuracy over the parameter-larger MLP-256 head while also slightly improving task-local accuracy. The frozen-policy pattern matches Fashion-MNIST: structural locking is available, while the plastic head is the better global seen-class evaluator under replay. This also clarifies the role of convolutional models in GrapNet: the experiment uses a conventional ResNet as the upstream visual module and applies child ownership after its output vector. Convolutional kernels can be viewed as local shared-weight relation aggregation between feature maps, but rewriting the visual encoder is unnecessary for CNN-backed use; the eye can remain a mature convolutional module while GrapNet supplies the editable downstream structure. Dense MLP heads and Transformer FFNs are still closer: by the mapping above, they are fully connected GrapNet special cases.

\subsection{Tabular Covariance Structure Editing}
A second programmability audit moves outside image classification. We use the Breast Cancer and Wine tabular datasets from scikit-learn \citep{pedregosa2011scikit}. Each standardized scalar feature enters through exactly one sensory GrapNode, so the input layer has 30 sensory nodes for Breast Cancer and 13 for Wine. The graph contains three hidden layers organized as shared-child execution families. During training, a covariance rule computes $|\mathrm{corr}(h_i,h'_j)|$ on training activations, removes the weakest child targets from each family, deletes hidden nodes that become zero-incoming, rebuilds the family backend, and continues training.

\begin{table}[t]
\centering
\scriptsize
\resizebox{\columnwidth}{!}{%
\begin{tabular}{lcccc}
\toprule
Dataset & Sensory & No edit acc. & Cov-edit acc. & Removed rel./nodes \\
\midrule
Breast Cancer & 30 & 96.49 $\pm$ 1.52 & 96.73 $\pm$ 1.20 & 22.9\% / 33.3\% \\
Wine & 13 & 98.89 $\pm$ 0.91 & 98.89 $\pm$ 0.91 & 32.1\% / 33.3\% \\
\bottomrule
\end{tabular}%
}
\caption{Non-image tabular covariance editing over five stratified splits. ``Sensory'' is the number of input GrapNodes and equals the number of scalar features. The last column reports deleted relations / hidden nodes; execution families are rebuilt after the edit.}
\label{tab:tabular}
\end{table}

Table~\ref{tab:tabular} shows structure editing as a first-class operation rather than a mask. On Breast Cancer, covariance editing removes 352 child relations and 32 hidden nodes per run while slightly increasing mean accuracy. On Wine, it removes 162 child relations and 18 hidden nodes per run while matching no-edit accuracy. The input contract is unchanged throughout: every sensory node consumes one feature dimension. These tabular studies are not meant to establish new UCI state of the art; they test whether the same child-owned substrate, family rebuilding, and physical deletion semantics work beyond image-feature heads.

\subsection{Execution and Structural Operations}
The trainable family backend matches the reference path with forward maximum absolute difference $2.24{\times}10^{-8}$ and maximum parameter-gradient difference $1.49{\times}10^{-8}$. In a three-seed dynamic training check, it reduces GrapNet+ER runtime by 1.75x and GrapNet+DER++ runtime by 1.85x relative to the reference child-owned path while preserving accuracy and forgetting. Timings that omit family execution are no-driver reference diagnostics; the dynamic accelerated path is the family backend. After topology stabilizes, a dense snapshot policy gives a 38.1x forward speedup on a full single-family transition, and a standalone C++ CPU relation backend gives a 2.68x speedup by moving accumulation out of Python.

The relation-attention policy audit gives the same guarantee for input-dependent routing. Before deletion, the policy matches a dense attention reference with zero measured forward, loss, child-weight-gradient, key-gradient, and query-gradient difference. After physically deleting five child relations, the attention mask updates from the graph and the same five differences remain zero. Gradients are nonzero for both child-owned $W_{alloc}$ coordinates and attention key/query parameters. In a teacher-student attention task, the same policy trains from 34.38\% to 92.19\% test accuracy. This experiment is included to show policy generality: family execution exploits repeated structural patterns, whereas attention execution uses input-dependent relation weights over the same canonical child-owned relation set.

Additional structural studies exercise the same invariant. An MLP-equivalent construction matches a dense PyTorch MLP to $1.2{\times}10^{-7}$ maximum absolute difference. On full-resolution 28x28 Fashion-MNIST organized as ten one-vs-rest online tasks, a child-owned grow-freeze model obtains 87.19\% final average accuracy with 0.00 measured forgetting, compared with 66.69\% and 26.18\% forgetting for a sequential MLP. A covariance-sifter rule operating over shared-child families physically reduces active relations from 3950 to 3750 on a Fashion-MNIST structural run while accuracy changes from 77.78\% to 78.17\%.

The supplement contains the released scripts, result JSON files, and command notes for the reported studies, including the Tiny Transformer FFN replacement audit. Dataset archives and ImageNet weights are not bundled; the artifact expects Fashion-MNIST and CIFAR-10 archives or the standard torchvision cache. The ResNet-feature study uses torchvision's ImageNet ResNet-18 weights when available and freezes the encoder before the online stream. This keeps the submission artifact lightweight while preserving the exact head-training scripts and aggregate statistics used in the paper.

\section{Discussion}
The experiments separate three variables that are often entangled in dynamic neural methods. Structural capacity controls how many editable hidden units enter the head. Replay controls sample memory and calibration. Freezing controls whether previously allocated hidden nodes become locked structural memory or remain plastic calibration resources. GrapNet makes these variables explicit because they are policies over child-owned nodes rather than implicit tensor surgery.

The main empirical pattern is consistent across Fashion-MNIST and CIFAR-10 ResNet features. Frozen structural memory is useful for isolating old computation, but global seen-class accuracy benefits from plastic old nodes trained under replay. This does not weaken the structural claim; it separates two operating modes. A deployment that values locked task modules can freeze them, while a class-incremental evaluator that must calibrate old and new logits can leave them plastic. The substrate exposes the choice instead of embedding it in a fixed tensor mask.

The structural localization audit is the clearest distinction from ordinary tensor masking. The selected object is a task-owned set of child relations in the architecture graph. The intervention physically removes those relations and produces a targeted computational effect. This is the operating model of a dynamic-architecture graph neural network: the graph is editable storage, causal program structure, and the source from which execution views are derived. Because the intervention is performed on child-reference/allocation-slot pairs, the same edit changes storage, gradients, and execution views together.

The runtime profile is also a consequence of the representation. Dense heads and static pretrained models are fastest when topology is fixed and no online structural program must be maintained. GrapNet pays time during online growth, deletion, freezing, and audit because those operations keep relations editable and visible to rules. Family execution is the native acceleration driver for this dynamic phase because it groups repeated child patterns without detaching gradients from live allocation coordinates. Dense snapshots and compiled relation kernels are stable-topology endpoints after edits settle. Thus the extra dynamic cost buys capabilities that a fixed pretrained head or a dense layer plus manual parameter surgery does not provide as a stable interface: runtime relation insertion/deletion, programmable freezing, local structural intervention, optimizer-aware parameter refresh, and execution-policy rebuilding from the same canonical graph.

\section{Scope}
GrapNet is evaluated here as an editable dynamic-architecture substrate behind a vector-producing parent interface. The parent can be raw vector input, a fully connected layer, a CNN encoder, a frozen ResNet feature extractor, an attention block, or a pooled/flattened transformer representation. Once a module emits a vector $z\in\mathbf{R}^D$, coordinate $z_i$ is consumed by sensory GrapNode $i$, and all coordinates can be processed in parallel. This is why the CIFAR-10 experiment uses a frozen ResNet-18 feature vector, while the tabular experiments use standardized scalar coordinates: both instantiate the same sensory contract. The same mathematics also allows internal substitution where the operator is an MLP: a dense head or Transformer FFN can be implemented as a fully connected GrapNet block. Operator-level versions of convolution and attention are compatible as well--convolution corresponds to local shared-weight aggregation across feature-map positions, and attention corresponds to input-dependent weighting over active relations--but the present experiments do not require replacing mature visual or token-mixing modules. The current evidence shows that child-owned heads can improve seen-class calibration under matched replay, preserve physical edit semantics across image and tabular inputs, and support multiple execution policies over the same canonical graph.

The reported results should therefore be read as evidence for a dynamic-architecture substrate rather than a claim that every backend is already throughput-optimal. No-driver reference timings measure the cost of preserving full editability in the simplest implementation. Family execution, attention execution, dense snapshots, and compiled relation kernels are the intended acceleration hierarchy: family for live dynamic training, dense snapshots and compiled kernels for stabilized throughput. The important systems property is that these backends are derived from one graph object: a structural edit updates the storage invariant first, and the execution policy follows.

\section{Conclusion}
GrapNet introduces dynamic-architecture graph neural networks in the graph-as-network setting. Any vector-producing parent module can feed sensory GrapNodes coordinate-wise, after which each current-layer node owns next-layer child references and the aligned allocation coordinates that implement outgoing computation. This invariant makes structural edits physical, policy-driven, and executable through multiple backends. Matched online studies on Split Fashion-MNIST and CIFAR-10 ResNet features show that an editable GrapNet replay head improves global seen-class accuracy over dense replay heads with comparable or larger parameter counts. Tabular covariance editing and structural localization show that the same substrate supports physical relation deletion and task-targeted causal interventions. Family and attention execution policies then turn the same child-owned graph into live matrix blocks or input-dependent routing kernels without detaching gradients from the canonical graph.

\end{document}